\documentclass[sn-vancouver,Numbered]{sn-jnl}

\usepackage{graphicx}%
\usepackage{multirow}%
\usepackage{amsmath,amssymb,amsfonts}%
\usepackage{amsthm}%
\usepackage{mathrsfs}%
\usepackage[title]{appendix}%
\usepackage{xcolor}%
\usepackage{textcomp}%
\usepackage{manyfoot}%
\usepackage{booktabs}%
\usepackage{algorithm}%
\usepackage{algorithmicx}%
\usepackage{algpseudocode}%
\usepackage{listings}%
\usepackage{natbib}
\usepackage{tabularx}
\usepackage{lmodern}

\begin{document}

\title[Article Title]{OTLP: Output thresholding using mixed integer linear programming}

\author*[1]{\fnm{Baran} \sur{Koseoglu}}\email{baran.koseoglu@wise.com}

\author[1]{\fnm{Luca} \sur{Traverso}}\email{luca.traverso@wise.com}

\author[1]{\fnm{Mohammed} \sur{Topiwalla}}\email{mohammed.topiwalla@wise.com}

\author[1]{\fnm{Egor} \sur{Kraev}}\email{egor.kraev@wise.com}

\author[2]{\fnm{Zoltan} \sur{Szopory}}\email{zoltan.szopory@wise.com}

\affil[1]{\orgdiv{Data Science}, \orgname{Wise Plc}, \orgaddress{\street{Shoreditch High St}, \city{London}, \postcode{E1 6JJ}, \country{UK}}}

\affil[2]{\orgdiv{Engineering}, \orgname{Wise Plc}, \orgaddress{\street{Shoreditch High St}, \city{London}, \postcode{E1 6JJ}, \country{UK}}}

\abstract{Output thresholding is the technique to search for the best threshold to be used during inference for any classifiers that can produce probability estimates on train and testing datasets. It is particularly useful in high imbalance classification problems where the default threshold is not able to refer to imbalance in class distributions and fail to give the best performance. This paper proposes OTLP, a thresholding framework using mixed integer linear programming which is model agnostic, can support different objective functions and different set of constraints for a diverse set of problems including both balanced and imbalanced classification problems. It is particularly useful in real world applications where the theoretical thresholding techniques are not able to address to product related requirements and complexity of the applications which utilize machine learning models. Through the use of Credit Card Fraud Detection Dataset, we evaluate the usefulness of the framework.}


\maketitle

\section{Introduction}\label{sec1}

Almost all classification methods such as XGBoost \cite{chen2016xgboost}, Random Forest \cite{breiman2001random}, Logistic Regression \cite{wang2019overview} are able to produce probability estimates. Output thresholding is a process to tune the decision threshold which is later used to assign class predictions based on a model's probability estimates for instances during inference \cite{sheng2006thresholding}. For binary classification tasks, instances with probability estimates higher than or equal to the threshold are assigned positives class, otherwise as negative which is depicted in Table~\ref{table:probability_estimates}. Adjusting the threshold is particularly important for imbalanced classification problems where the train datasets have a smaller number of samples in the minority classes compared to the other classes. Output thresholding is one of the methods to address class imbalance problem \cite{esposito2021ghost}. Since the distribution of classes is skewed and probability estimates often favor the majority class, using a default classification threshold of 0.5 may not be the most effective approach for such problems \cite{leevy2023threshold}. Therefore it is essential to perform a search for the threshold to use during inference. Output thresholding is also considered to address class imbalance problem for convolutional neural networks~\cite{buda2018systematic}.

\begin{table}[htbp]
\centering
\begin{tabularx}{\textwidth}{|XXX|} 
 \hline
 \textbf{Probability Estimate} & \textbf{Threshold} & \textbf{Label} \\ [0.5ex] 
 \hline
 0.4 & 0.5 & 0 \\ 
 0.56 & 0.5 & 1 \\
 0.61 & 0.5 & 1 \\
 0.14 & 0.5 & 0 \\
 0.88 & 0.5 & 1 \\ [1ex] 
 \hline
\end{tabularx}
\caption{A classifier provides probability estimates for instances which is recorded under Probability Estimate column. This is know as soft classification. Then the instance with the predicted probability higher than or equal to the threshold is assigned 1 and 0 otherwise. }
\label{table:probability_estimates}
\end{table}

\section{Related Work}\label{sec2}

\begin{figure}[tb]
 \centering 
 \includegraphics[scale=0.6]{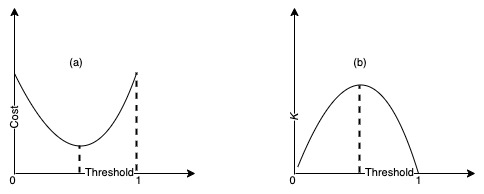}
 \caption{(a) shows a curve of mislassification cost with increasing threshold values in range $[0,1]$ whereas (b) shows a curve of Cohen's kappa metric with increasing threshold values in range. The objective function and related metrics is subject to change according to the characteristics of the dataset and the problem at hand.}
 \label{fig:objective_function_example}
\end{figure}

As an optimization procedure output thresholding finds a global minimum/maximum to an objective function. Objective function is a function of thresholds. Metrics used in the objective function can differ according to application and the characteristics of the dataset. Balanced metrics such as the Cohen's kappa or Matthews correlation coefficient are amongst the most suitable metrics to measure performance of imbalanced datasets \cite{fatourechi2008comparison} and are likely to be used as objective functions for imbalanced datasets. While performing the search for the optimal threshold, a confusion matrix is generated for every threshold candidate based on the classes assigned on the instances and metrics used in the objective function are calculated. Output thresholding can be performed during training on validation datasets or on samples of training datasets. It can also be performed as a post-processing step after training the model. The decision to select which dataset to use for searching the optimal threshold can introduce a bias \cite{sheng2006thresholding}. We can formulate the procedure as follows:
\begin{equation*}
\begin{aligned}
& \underset{t}{\text{minimize}}
& & C(T) \\
& \text{subject to}
& & 0<=t<=1 \\
\end{aligned}
\end{equation*} 

where function $C(T)$ represents an objective function of the threshold variable $T$. The curve of this function can be obtained after computing confusion matrix and the related metrics for each threshold in the range $0<=t<=1$. Figure~\ref{fig:objective_function_example} shows different objective functions with increasing threshold values in range $[0,1]$. We can see that thresholding selects the threshold minimizing the cost in one while it selects the threshold maximizing Cohen's kappa in the other.

There exist several studies which are aimed at finding the optimal decision threshold. Esposito et al.~\cite{esposito2021ghost} proposes to optimize the decision threshold according to Cohen's kappa metric which is the selected objective function in this case. Although the methodology proposed is model agnostic and doesn't require retraining of the machine learning model, the optimization algorithm doesn't support any constraints and evaluation is performed very specifically to compare performance of different methodologies including output thresholding on class imbalance problem. Furthermore the optimal thresholds can be biased as thresholds are evaluated on samples of training sets. Similar to ~\cite{movahedi2010limitations} Zou et al. ~\cite{zou2016finding} suggests that receiver operating characteristic (ROC) curve can evaluate the performance of a classifier by giving equal weight to both majority and minority classes. However, they note that this method is not ideal for imbalanced datasets because the performance on the minority classes is more crucial. F-score thus can be used to find the optimal threshold for imbalanced datasets which prioritizes the performance of the classifier on the minority classes. Both of these studies focus only on class imbalance problem. Additionally, the algorithm suggested by ~\cite{zou2016finding} is specifically designed to work with the F-score metric, as mentioned in their study.

Leevy et al.~\cite{leevy2023threshold} evaluated how random undersampling affects threshold optimization and found that this technique generally reduces the performance of machine learning models. Although the study tested various metrics and involved 1 type of constraint for threshold optimization, their algorithm doesn't support more complex constraints. Some of their conclusions are also based on Geometric mean of true positive and true negative rate which may be biased to differences in distribution of class labels for imbalanced datasets~\cite{movahedi2010limitations}. In this study, we
challenge the problem of output thresholding utilizing mixed integer linear programming.  In doing so, we present OTLP, a thresholding framework that aims to find the optimized threshold, which is model agnostic and can support a diverse set of constraints and objective functions for both balanced and imbalanced datasets. We evaluate our proposed solution by testing it with a Credit Card Fraud Detection Dataset~\cite{datacite}. To
support our claims on usefulness of the framework, we present several experiments which differ from one another by either classifier type, class ratio in the dataset, set of constraints or objective function. In summary, our contributions comprise a thresholding framework which:\begin{itemize}
\itemsep 0em 
  \item Uses mixed integer linear programming on output thresholding,
  \item Supports basic as well as complex constraints,
  \item Can be used for balanced and imbalanced classification problems,
  \item Allows to use customized objective functions, and
  \item Is model agnostic.
   
\end{itemize}
The rest of this paper is structured as follows: Section~\ref{sec3} describes the methods used in OTLP; Section~\ref{sec4} covers the design of the experiments and their outcomes; Section~\ref{sec5} highlights the main findings of this study and offers recommendations for further research.

\begin{figure}[tb]
 \centering 
\includegraphics[width=\linewidth]{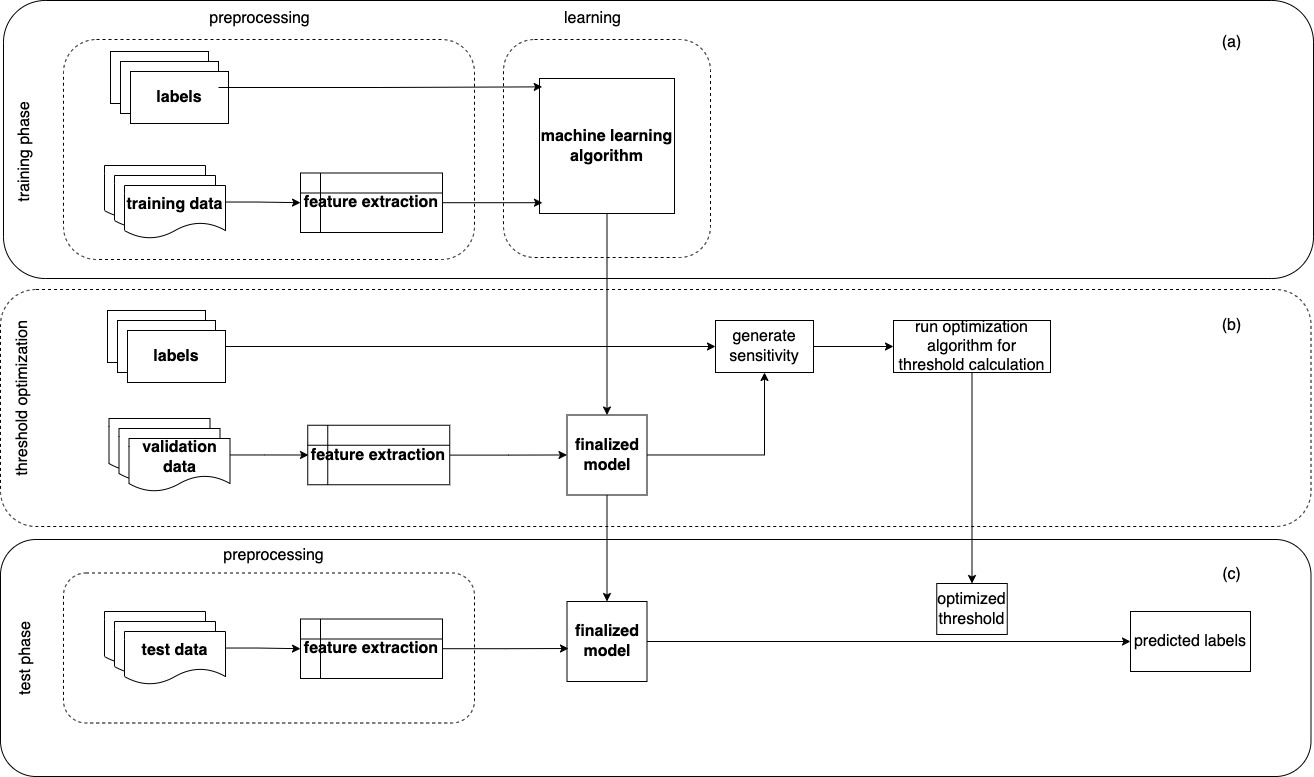} 
 \caption{shows in which step the threshold optimization framework is run with respect to a classical machine learning model training flow. A machine learning model is trained in step (a). Sensitivity is generated on validation dataset using the trained model in step (a) which is further used by OTLP to select the optimized threshold in step (b). The threshold is later used to assign classes to the instances in the test dataset in step (c).}
 \label{fig:OTLP_framework}
\end{figure}

\section{Methodology}\label{sec3}

OTLP uses mixed integer linear programming to find the optimized thresholds. After a classifier is trained on the training set, we generate confusion matrix for each threshold in a specified range of thresholds using validation dataset. Confusion matrix is further processed to assess the performance of the classifier at the specified threshold and is logged to a tabular structured sheet named sensitivity. Each row in the sensitivity represents a threshold while columns represent metrics like true positive, false positive, true negative, false negative, precision, recall and F-score. Any application specific objective function value can be added to the column list. Table~\ref{table:sensitivity_example} shows an example for sensitivity of the specified structure.

\begin{table}[htbp]
\centering
\begin{tabularx}{\textwidth}{|c|c|c|c|c|X|X|c|c|}  
 \hline
 \textbf{Threshold} & \textbf{TP} & \textbf{FP} & \textbf{TN} & \textbf{FN} & \textbf{Precision} & \textbf{Recall} & \textbf{F1 Score} & \textbf{Total Cost} \\ [0.5ex] 
 \hline
 0.05 & 48 & 15 & 85 & 2 & 0.762 & 0.960 & 0.849 & \$160 \\ 
 0.1 & 45 & 10 & 90 & 5 & 0.818 & 0.900 & 0.857 & \$125 \\ 
 0.35 & 40 & 5 & 95 & 10 & 0.889 & 0.800 & 0.842 & \$100 \\ 
 0.55 & 35 & 3 & 97 & 15 & 0.921 & 0.700 & 0.796 & \$105 \\ 
 0.71 & 30 & 2 & 98 & 20 & 0.937 & 0.600 & 0.731 & \$120 \\  [1ex] 
 \hline
\end{tabularx}
\caption{Example table with threshold, true positives (TP), false positives (FP), true negatives (TN), false negatives (FN), precision, recall, F1 score, and total cost.}
\label{table:sensitivity_example}
\end{table}

 After sensitivity is generated, mixed integer linear programming is run to find optimized threshold which minimizes/maximizes an objective function with a specified list of constraints. This threshold is further used for test dataset during inference to assign class labels to the instances. Figure~\ref{fig:OTLP_framework} shows how OTLP can be used as part of a classical machine learning model training flow. 

 In order to select the optimized threshold using mixed integer linear programming, decision variables need to be defined. We can define a linear programming problem minimizing a loss function to select the optimized threshold as follows \begin{equation*}
\begin{aligned}
& \underset{t}{\text{minimize}}
& & \sum_{i=1}^{N}{a_{i}*t_{i}} \\
& \text{subject to}
& & t_{i} \in \{0,1\}, \quad i=1,2,\ldots,N \\
& & & t_{1}+t_{2} \ldots = 1 \\
\end{aligned}
\end{equation*} 

where $a_{i}$ is the scalar representing loss for the respective threshold $t_{i}$. By defining decision variables for thresholds, output thresholding becomes an optimization problem which selects the threshold minimizing the objective function. Leveraging mixed integer linear programming to find the optimized threshold has a major advantage. Constraints of different level of complexities can be easily added to the optimization problem. 

\subsection{Basic Constraints}\label{subsec3.1}

OTLP supports adding basic constraints to the optimization problem. These sets of constraints are specific conditions which restrict the values of thresholds within a particular range. An example would be optimizing threshold for a machine learning model where we want to limit the sum of true positives and false positives. This could be due to the fact we want to limit number of instances we want to assign positive class to. The problem then becomes \begin{equation*}
\begin{aligned}
& \underset{t}{\text{minimize}}
& & \sum_{i=1}^{N}{a_{i}*t_{i}} \\
& \text{subject to}
& & t_{i} \in \{0,1\}, \quad i=1,2,\ldots,N \\
& & & t_{1}+t_{2} \ldots = 1 \\
& & & b_{1}*t_{1}+b_{2}*t_{2} \ldots <= C \\
\end{aligned}
\end{equation*} 

where $a_{i}$ is the scalar representing loss, $b_{i}$ is the sum of true positive and false positive for the respective threshold $t_{i}$ whereas $C$ is the total number of instances we want to assign positive class to.

\subsection{Complex Constraints}\label{subsec3.2}

In real-world applications, datasets often exhibit complex structures where instances are not uniformly distributed across all features. In many cases, it is beneficial to consider subsets of the dataset, referred to as subspaces, and optimize thresholds independently within each subspace. Let $\mathcal{X}$ denote the space where all instances belong to. Each of the instances in the original space $\mathcal{X}$ is characterized by features $\{{f_{1},f_{2},\ldots,f_{n}}\}$ where $f_{j}$ represents the $j^{th}$ feature used in the dataset. If we create two subspace $\mathcal{X}_1 \quad \text{and} \quad \mathcal{X}_2$ characterized by a specific feature $f_{j}$ in the original dataset, then we can assign every instance to one of these subspaces. The threshold optimization problem can be addressed independently within each subspace but these optimization problems can be subject to global constraints, conditions that apply universally across these subspaces, such as total number of true positives and false positives in the original space. OTLP supports global constraints as well. The problem now becomes \begin{equation*}
\begin{aligned}
& \underset{t}{\text{minimize}}
& & \sum_{i=1}^{2} \sum_{j=1}^{N}{a_{ij}*t_{ij}} \\
& \text{subject to}
& & t_{ij} \in \{0,1\}, \quad i=1,2,  \quad j=1,2,\ldots,N \\
& & & t_{11}+t_{12},\ldots,+t_{1N} = 1 \\
& & & t_{21}+t_{22},\ldots,+t_{2N} = 1 \\
& & & b_{11}*t_{11}+b_{12}*t_{12},  \ldots ,+b_{2N}*t_{2N}<= C \\
\end{aligned}
\end{equation*} 

where $a_{ij}$ is the scalar representing loss, $b_{ij}$ is the sum of true positive and false positive for the respective threshold $t_{ij}$ in the space $i$ whereas $C$ is the total number of instances we want to assign positive class to. OTLP selects one and only one threshold for the first subspace while it selects another threshold for the second subspace. OTLP has a global constraint on the total number of true positives and false positives across the subspaces.

\section{Experiments}\label{sec4}
Experiments were run on a machine with 6-Core Intel Core i7 processor and 32GB memory. For every experiment, we changed one of the configurations in the following list: classifier type, class ratio in the dataset, objective function and list of constraints we apply to the optimization algorithm as shown in table~\ref{table:experiment_design}. Classifier type represents the classification method used in the experiments whereas class ratio denotes the proportion of positive to negative classes in the training set, with a 1:1 ratio indicating an equal number of positive and negative classes. Objective function is the function OTLP optimizes and list of constraints comprises the constraints that restrict the search space for the optimized threshold. This setup is configured to show that OTLP is model agnostic, supports basic as well as complex constraints, can be used for both balanced and imbalanced classification problems and can optimize a diverse set of objective functions including F-scores. We don't optimize hyperparameters used in the classifiers as the objective of the experiments is not to find the most performant setting. 

For XGBoost classifier, we used xgboost library \cite{xgboost} with following hyperparameters: 0.8 as colsample bytree, 10 as max depth, 0.1 as learning rate, 100 as n estimators, 0.8 as subsample, binary:logistic as objective and gbtree as booster. For Random Forest classifier, we used Scikit-learn library \cite{pedregosa2011scikit} and the following hyperparameters: gini as criterion, sqrt as max features, 0 as ccp alpha, 0 as min impurity decrease, 1 as min samples leaf, 2 as min samples split, 0 as min weight fraction leaf, 100 as n estimators, true as bootstrap. For all of the experiments, we used the same dataset which we split into three as train, validation and test dataset with respective ratios 0.70, 0.20, 0.10. We used the train dataset for training the classifier, used validation dataset to find optimal thresholds and tested them on the test dataset. One can apply cross validation while finding optimized thresholds but according to our experiments running OTLP on single validation set yielded similar results compared to cross validation.

\begin{table}[htbp]
\centering
\begin{tabularx}{\textwidth}{|c|c|c|c|X|} 
 \hline
 \textbf{Experiment} & \textbf{Classifier Type} & \textbf{Class Ratio} & \textbf{Objective Function} & \textbf{Constraints} \\ [0.5ex] 
 \hline
 1 & XGBoost & Original & F1-score & Local \\ 
 2 & XGBoost & 1:1 & F1-score & Local \\
 3 & Random Forest & Original & F1-score & Local\\
 4 & XGBoost & Original & F1-score & Local \\
 5 & XGBoost & Original & F2-score & Local \\
 6 & XGBoost & Original & Loss & Local and Global \\[1ex] 
 \hline
\end{tabularx}
\caption{Experiments differ from each other in one of the configurations in the following
list: classifier type, class ratio, objective function and list of constraints.}
\label{table:experiment_design}
\end{table}

\begin{figure}[tb]
 \centering 
 \includegraphics[scale=0.4]{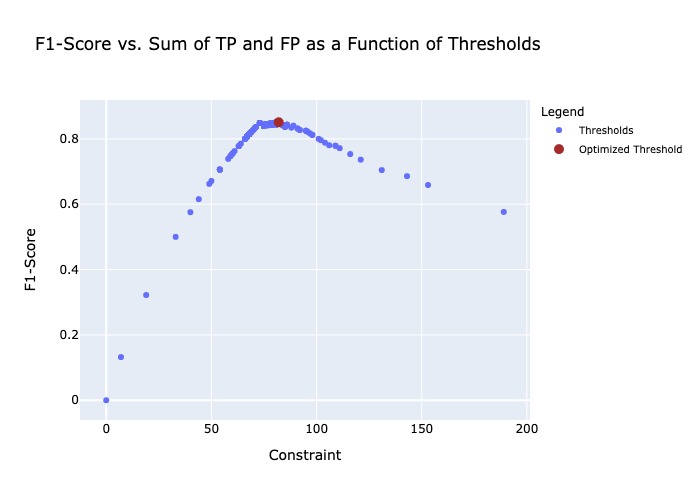}
 \caption{OTLP selects the optimized threshold as the global maxima of the objective function within the specified constraint list.}
 \label{fig:4_1_results_fig}
\end{figure}

\subsection{Results for the original class ratio}\label{sec4.1}
We trained an XGBoost classifier on the dataset with original class ratio. The metric used in the objective function was F1-score. We limited the total number of true positives and false positives to 200 as a constraint. Figure~\ref{fig:4_1_results_fig} shows that OTLP is able to find optimized threshold taking into consideration of the constraint for the validation set.

\begin{table}[htbp]
\centering
\begin{tabularx}{\textwidth}{|c|c|c|c|c|X|X|X|} 
 \hline
 \textbf{Threshold} & \textbf{TP} & \textbf{FP} & \textbf{TN} & \textbf{FN} & \textbf{Precision} & \textbf{Recall} & \textbf{F1-score} \\ [0.5ex] 
 \hline
 0.2 & 41 & 5 & 28143 & 8 & 0.89 & 0.84 & \textbf{0.86}\\ 
 0.5 & 37 & 3 & 28145 & 12 & 0.92 & 0.75 & \textbf{0.83} \\ [1ex] 
 \hline
\end{tabularx}
\caption{Threshold selected by OTLP framework performs better than the default threshold in same constrained space.}
\label{table:4_1_results_tab}
\end{table}

\begin{table}[htbp]
\centering
\begin{tabularx}{\textwidth}{|c|c|c|c|c|X|X|X|} 
 \hline
 \textbf{Threshold} & \textbf{TP} & \textbf{FP} & \textbf{TN} & \textbf{FN} & \textbf{Precision} & \textbf{Recall} & \textbf{F1-score} \\ [0.5ex] 
 \hline
 0.2 & 41 & 5 & 28143 & 8 & 0.89 & 0.84 & \textbf{0.863}\\ 
 0.075 & 41 & 6 & 28142 & 8 & 0.87 & 0.84 & \textbf{0.854} \\
 0.27 & 40 & 5 & 28143 & 9 & 0.89 & 0.82 & \textbf{0.851} \\[1ex] 
 \hline
\end{tabularx}
\caption{Thresholds and respective metrics sorted by descending F1-score in the test dataset.}
\label{table:4_1_results_thresholds_tab}
\end{table}

Table~\ref{table:4_1_results_tab} shows that 0.2 threshold selected by OTLP framework performs better than default 0.5 threshold in the test dataset. On the other hand, table~\ref{table:4_1_results_thresholds_tab} shows the top 3 entries of the thresholds and the respective metrics sorted by descending F1-score in the test dataset. It can be confirmed that 0.2 threshold selected by OTLP is the optimized threshold in the test dataset.

\begin{figure}[tb]
 \centering 
 \includegraphics[scale=0.4]{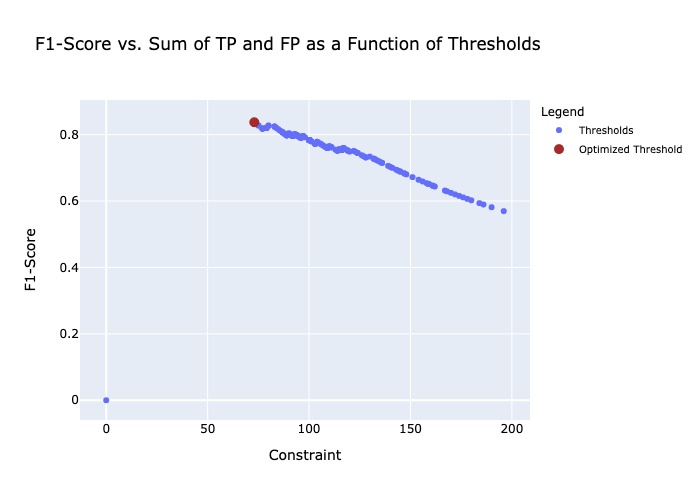}
 \caption{OTLP selects the optimized threshold as the global maxima of the objective function within the specified constraint list in the validation dataset. The train dataset used to train the model was balanced with 1:1 class ratio.}
 \label{fig:4_2_results_fig}
\end{figure}

\subsection{Results for the balanced class ratio}\label{sec4.2}
We trained an XGBoost classifier on the dataset with 1:1 class ratio between positive and negative classes. Synthetic Minority Oversampling Technique (SMOTE) \cite{chawla2002smote} was used to generate synthetic samples. The metric used in the objective function was F1-score. We limited the total number of true positives and false positives to 200 as a constraint.  Figure~\ref{fig:4_2_results_fig} shows that OTLP is able to find optimized threshold for the balanced dataset.

\begin{table}[htbp]
\centering
\begin{tabularx}{\textwidth}{|c|c|c|c|c|X|X|X|} 
 \hline
 \textbf{Threshold} & \textbf{TP} & \textbf{FP} & \textbf{TN} & \textbf{FN} & \textbf{Precision} & \textbf{Recall} & \textbf{F1-score}  \\ [0.5ex] 
 \hline
 0.99 & 38 & 2 & 28146 & 11 & 0.95 & 0.77 & \textbf{0.85} \\ 
 0.5 & 42 & 17 & 28131 & 7 & 0.71 & 0.86 & \textbf{0.78}\\ [1ex] 
 \hline
\end{tabularx}
\caption{Threshold selected by OTLP framework performs better than the default threshold in same constrained space.}
\label{table:4_2_results_tab}
\end{table}

\begin{table}[htbp]
\centering
\begin{tabularx}{\textwidth}{|c|c|c|c|c|X|X|X|} 
 \hline
 \textbf{Threshold} & \textbf{TP} & \textbf{FP} & \textbf{TN} & \textbf{FN} & \textbf{Precision} & \textbf{Recall} & \textbf{F1-score}  \\ [0.5ex] 
 \hline
 0.99 & 38 & 2 & 28146 & 11 & 0.95 & 0.77 & \textbf{0.854} \\ 
 0.88 & 40 & 5 & 28143 & 9 & 0.89 & 0.82 & \textbf{0.851}\\
 0.905 & 39 & 4 & 28144 & 10 & 0.91 & 0.79 & \textbf{0.848}\\[1ex] 
 \hline
\end{tabularx}
\caption{Thresholds and respective metrics sorted by descending F1-score in the test dataset.}
\label{table:4_2_results_thresholds_tab}
\end{table}

Table~\ref{table:4_2_results_tab} shows that 0.99 threshold selected by OTLP framework performs better than default 0.5 threshold in the test dataset. OTLP framework works for balanced datasets as well as imbalanced datasets. Table~\ref{table:4_2_results_thresholds_tab} shows the top 3 entries of the thresholds and the respective metrics sorted by descending F1-score in the test dataset. It can be confirmed that 0.99 threshold selected by OTLP is the optimized threshold for the test set.

\begin{figure}[tb]
 \centering 
 \includegraphics[scale=0.4]{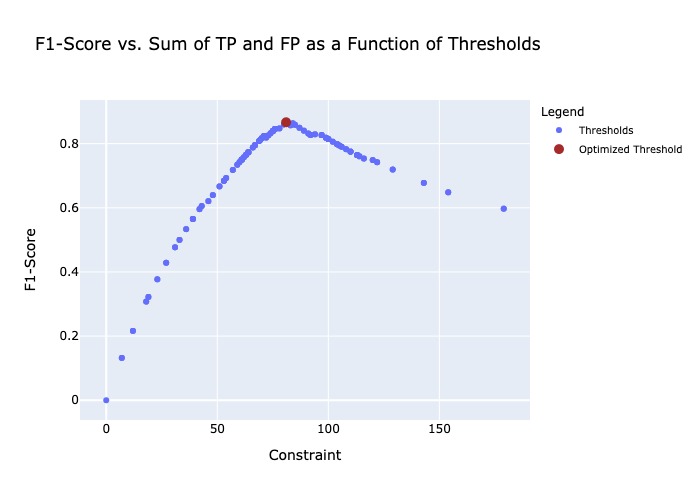}
 \caption{OTLP selects the optimized threshold as the global maxima of the objective function within the specified constraint list in the validation dataset.}
 \label{fig:4_3_results_fig}
\end{figure}

\subsection{Results for Random Forest algorithm}\label{sec4.3}
In order to show that OTLP framework is model agnostic, we trained a Random Forest classifier on the dataset with  with original class ratio different from the previous experiments. The metric used in the objective function was F1-score. We limited the total number of true positives and false positives to 200 as a constraint. Figure~\ref{fig:4_3_results_fig} shows that OTLP is able to find optimized threshold when a different classifier type is used during model training.

\begin{table}[htbp]
\centering
\begin{tabularx}{\textwidth}{|c|c|c|c|c|X|X|X|} 
 \hline
 \textbf{Threshold} & \textbf{TP} & \textbf{FP} & \textbf{TN} & \textbf{FN} & \textbf{Precision} & \textbf{Recall} & \textbf{F1-score} \\ [0.5ex] 
 \hline
 0.4 & 40 & 3 & 28145 & 9 & 0.93 & 0.82 & \textbf{0.87} \\ 
 0.5 & 39 & 3 & 28145 & 10 & 0.93 & 0.79 & \textbf{0.86} \\ [1ex] 
 \hline
\end{tabularx}
\caption{Threshold selected by OTLP framework performs better than the default threshold in same constrained space.}
\label{table:4_3_results_tab}
\end{table}

\begin{table}[htbp]
\centering
\begin{tabularx}{\textwidth}{|c|c|c|c|c|X|X|X|} 
 \hline
 \textbf{Threshold} & \textbf{TP} & \textbf{FP} & \textbf{TN} & \textbf{FN} & \textbf{Precision} & \textbf{Recall} & \textbf{F1-score} \\ [0.5ex] 
 \hline
 0.4 & 40 & 3 & 28145 & 9 & 0.93 & 0.82 & \textbf{0.869} \\ 
 0.37 & 40 & 4 & 28144 & 9 & 0.91 & 0.82 & \textbf{0.860} \\ 
 0.43 & 39 & 3 & 28145 & 10 & 0.93 & 0.79 & \textbf{0.857} \\ [1ex] 
 \hline
\end{tabularx}
\caption{Thresholds and respective metrics sorted by descending F1-score in the test dataset.}
\label{table:4_3_results_thresholds_tab}
\end{table}

Table~\ref{table:4_3_results_tab} shows that 0.4 threshold selected by OTLP framework performs better than default 0.5 threshold for test dataset. OTLP framework works for different type of classifiers. Table~\ref{table:4_3_results_thresholds_tab} shows the top 3 entries of the thresholds and the respective metrics sorted by descending F1-score in the test dataset. It can be confirmed that 0.4 threshold selected by OTLP is the optimized threshold for the test set.

\begin{figure}[tb]
 \centering 
 \includegraphics[scale=0.4]{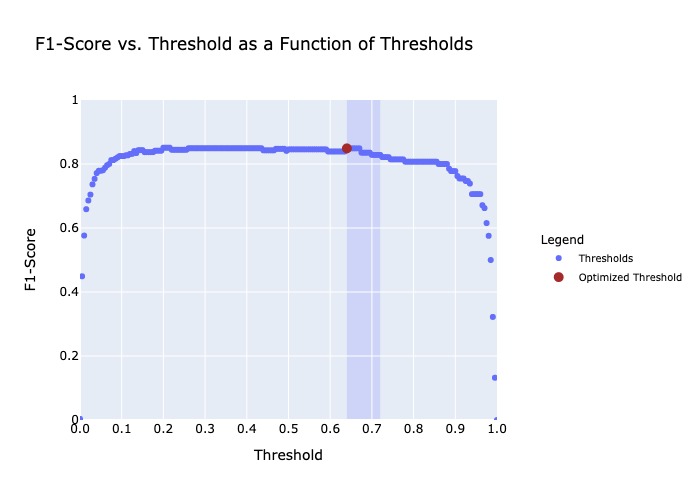}
 \caption{OTLP selects the optimized threshold as the global maxima of the objective function within the threshold range satisfying the constraints. Thresholds satisfying constraints are shown in the shaded area. OTLP optimizes objective function among those thresholds in the validation dataset.}
 \label{fig:4_4_results_fig}
\end{figure}

\subsection{Results for a different set of constraints}\label{sec4.4}
OTLP framework can support diverse set of constraints. Experiments run so far limited total number of false positives and true positives. We replaced the previous constraint on total number of true positives and false positives with constraint on precision and recall and trained an XGBoost classifier on the dataset with original class ratio. Precision is set to be at least 0.98 whereas recall is set to be at least 0.70. The metric used in the objective function was F1-score. Figure~\ref{fig:4_4_results_fig} shows that OTLP is able to find optimized threshold for this setting. 

Unlike the previous experiments, there is no threshold satisfying the specified precision and recall constraint for this setting in the test dataset. Table~\ref{table:4_4_results_thresholds_tab} shows the top 3 entries of the thresholds and the respective metrics sorted by descending F1-score having precision greater than or equal to 0.92 and recall greater than or equal to 0.70 in the test dataset. 0.64 threshold selected by OTLP using the validation dataset performs in top 3 thresholds maximizing F1-score.

\begin{figure}[tb]
 \centering 
 \includegraphics[scale=0.4]{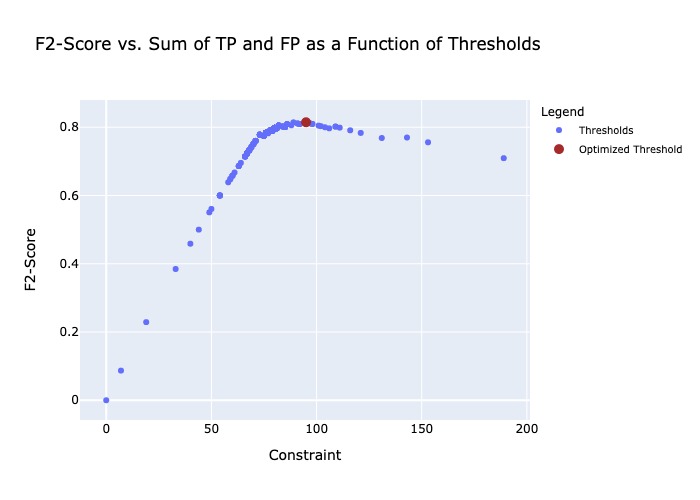}
 \caption{OTLP selects the optimized threshold as the global maxima of the objective function within the threshold range satisfying the constraints in the validation dataset.}
 \label{fig:4_5_results_fig}
\end{figure}

\begin{table}[htbp]
\centering
\begin{tabularx}{\textwidth}{|c|c|c|c|c|X|X|X|} 
 \hline
 \textbf{Threshold} & \textbf{TP} & \textbf{FP} & \textbf{TN} & \textbf{FN} & \textbf{Precision} & \textbf{Recall} & \textbf{F1-score} \\ [0.5ex] 
 \hline
 0.43 & 38 & 3 & 28145 & 11 & 0.93 & 0.77 & \textbf{0.84} \\ 
 0.505 & 37 & 3 & 28145 & 12 & 0.92 & 0.75 & \textbf{0.83} \\ 
 0.64 & 36 & 3 & 28145 & 13 & 0.92 & 0.73 & \textbf{0.82} \\ [1ex] 
 \hline
\end{tabularx}
\caption{Thresholds and respective metrics sorted by descending F1-score in the test dataset.}
\label{table:4_4_results_thresholds_tab}
\end{table}

\subsection{Results for a different objective function}\label{sec4.5}
OTLP is able to optimize threshold for customized objective functions. We used F2-score as metric in the objective function and trained an XGBoost classifier on the dataset with original class ratio. We limited the total number of true positives and false positives to 200 as a constraint. Figure~\ref{fig:4_5_results_fig} shows that OTLP is able to find optimized threshold for customized objective function within the specified constraint in the validation dataset.

\begin{table}[htbp]
\centering
\begin{tabularx}{\textwidth}{|c|c|c|c|c|X|X|X|} 
 \hline
 \textbf{Threshold} & \textbf{TP} & \textbf{FP} & \textbf{TN} & \textbf{FN} & \textbf{Precision} & \textbf{Recall} & \textbf{F2-score} \\ [0.5ex] 
 \hline
 0.1 & 41 & 6 & 28142 & 8 & 0.87 & 0.84 & \textbf{0.84} \\ 
 0.5 & 37 & 3 & 28145 & 12 & 0.92 & 0.75 & \textbf{0.78} \\ [1ex] 
 \hline
\end{tabularx}
\caption{Threshold selected by OTLP framework performs better than the default threshold in same constrained space.}
\label{table:4_5_results_tab}
\end{table}

\begin{table}[htbp]
\centering
\begin{tabularx}{\textwidth}{|c|c|c|c|c|X|X|X|} 
 \hline
 \textbf{Threshold} & \textbf{TP} & \textbf{FP} & \textbf{TN} & \textbf{FN} & \textbf{Precision} & \textbf{Recall} & \textbf{F2-score} \\ [0.5ex] 
 \hline
 0.245 & 41 & 5 & 28143 & 8 & 0.89 & 0.84 & \textbf{0.847} \\ 
 0.1 & 41 & 6 & 28142 & 8 & 0.87 & 0.84 & \textbf{0.844} \\ 
 0.055 & 41 & 7 & 28141 & 8 & 0.85 & 0.84 & \textbf{0.840} \\ [1ex] 
 \hline
\end{tabularx}
\caption{Thresholds and respective metrics sorted by descending F2-score in the test dataset.}
\label{table:4_5_results_thresholds_tab}
\end{table}

Table~\ref{table:4_5_results_tab} shows that 0.1 threshold selected by OTLP framework performs better than default 0.5 threshold in the test dataset. Table~\ref{table:4_5_results_thresholds_tab} shows the top 3 entries of the thresholds and the respective metrics sorted by descending F2-score in the test dataset. It
can be confirmed that 0.1 threshold selected by OTLP is one of the thresholds having the highest F2-score for
the test dataset.

\subsection{Results for complex set of constraints}\label{sec4.6}
In real-world applications, datasets frequently demonstrate complex structures where instances aren't evenly spread across all features. Under such circumstances, it proves advantageous to examine subsets of the dataset and create subspaces. Optimizing thresholds within each subspace, as previously discussed, may be indeed the objective for such datasets. Although we can apply threshold optimization to these subspaces independently, they may have dependent constraints such as sum of true positives and false positives. OTLP is able to find optimized thresholds for such problems and supports local as well as global constraints. We trained an XGBoost classifier on the train dataset with original class ratio. We then created two subspaces in the original validation and test datasets by assigning a class to instances which are used to create two different sensitivies. Assigning a class to instances created two subspaces in both validation and test datasets which are mutually exclusive. We then run OTLP framework on the concatenated sensitivities to find optimal thresholds for these two subspaces in the validation dataset. The metric used in the objective function was the loss associated with false negatives. We limited the total number of true positives and false positives to 200 as a constraint which is a global constraint affecting both of the subspaces. We also constrainted thresholds to have at least 0.5 precision which is a local constraint. 

\begin{figure}[tb]
 \centering
 \includegraphics[scale=0.4]{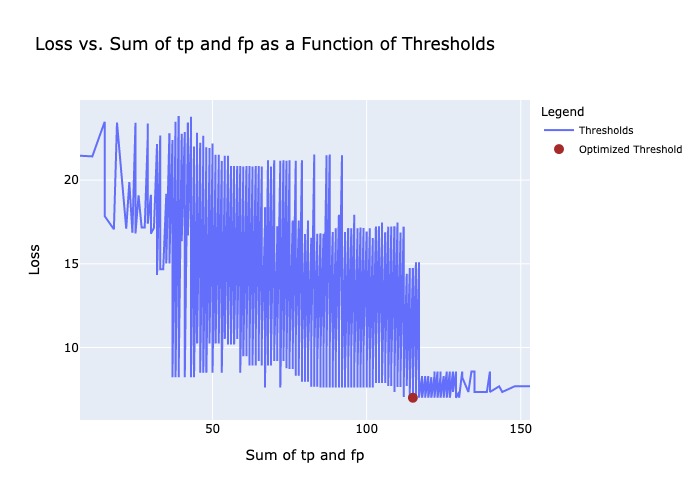}
 \caption{OTLP selects the optimized threshold as the global minima of the objective function within the threshold range satisfying both global and local constraints.}
 \label{fig:4_6_results_fig}
\end{figure}

Table~\ref{table:4_6_results_tab} shows the respective thresholds selected by OTLP on the validation set. In order to show thresholds suggested by OTLP are optimal, we iterated over all possible tuples of thresholds for the two subspaces which satisfy the constraints. Figure~\ref{fig:4_6_results_fig} shows that thresholds selected by OTLP for the two subspaces are the optimal thresholds minimizing the loss function among all the other threshold tuples with total loss 6.99. The same set of thresholds proposed by OTLP are also optimal in the test dataset. Figure~\ref{fig:4_6_results_fig} also illustrates that the optimization landscape is not smooth, a condition where traditional optimization methods may not perform effectively. This observation further justifies the application of mixed integer linear programming to determine optimized thresholds in such a challenging landscape. 

\begin{table}[htbp]
\centering
\begin{tabularx}{\textwidth}{|c|c|c|c|c|c|c|X|X|} 
 \hline
 \textbf{Subspace} & \textbf{Threshold} & \textbf{TP} & \textbf{FP} & \textbf{TN} & \textbf{FN} & \textbf{Precision} & \textbf{Recall} & \textbf{Loss} \\ [0.5ex] 
 \hline
 1 & 0.11 & 45 & 4 & 28566 & 8 & 0.92 & 0.85 & 5.16 \\ 
 2 & 0.02 & 36 & 30 & 28547 & 10 & 0.54 & 0.78 & 1.83 \\ [1ex] 
 \hline
\end{tabularx}
\caption{Threshold selected by OTLP framework for the respective subspaces in the validation set.}
\label{table:4_6_results_tab}
\end{table}

\section{Conclusion}\label{sec5}
In our research, we proposed OTLP, an output threshold framework using mixed integer linear programming, which is model agnostic, can support different objective functions and different set of constraints for a diverse set of problems including both balanced and imbalanced classification problems. OTLP is particularly useful for real world applications where the theoretical thresholding techniques are not able to address to complex data structures and complexities of the business applications which require to divide the problem into several sub-problems and to find optimized thresholds for each subproblem. In order to show that OTLP finds optimal thresholds in such complex circumstances, we run several experiments and found optimized thresholds that other output thresholding algorithms fail to achieve. Thresholds proposed by OTLP framework were within the top 3 optimal thresholds for the test dataset which suggests that OTLP can be used as an interim step in a classical machine learning model training flow to select optimized threshold which is later used to assign class predictions for instances during inference. Experiments show that OTLP as an output thresholding framework proposes thresholds which perform better than the default 0.5 threshold. For future work, we will seek to answer if OTLP can scale as the dataset size gets bigger.

\end{document}